\documentclass{article}
\usepackage{spconf,amsmath,epsfig}
\usepackage{algorithmic} % new added
\usepackage{algorithm}
\usepackage{amsmath}
\usepackage{amsfonts}
\usepackage{mathtools}
\usepackage{multirow}

\begin{document}\sloppy

% Example definitions.
% --------------------
\def\x{{\mathbf x}}
\def\L{{\cal L}}

% Title.
% ------
\title{Post-Processing of Word Representations via Variance Normalization and Dynamic Embedding}
%
% Single address.
% ---------------

\name{Bin Wang\textsuperscript{1}, Fenxiao Chen\textsuperscript{1}, Angela Wang\textsuperscript{2}, C.-C. Jay Kuo\textsuperscript{1}}
\address{University of Southern California\textsuperscript{1}, University of California, Berkeley\textsuperscript{2}}
\maketitle
\begin{abstract}
Language processing becomes more and more important in multimedia processing. Although embedded vector representations of words offer impressive
performance on many natural language processing (NLP) applications, the
information of ordered input sequences is lost to some extent if only
context-based samples are used in the training. For further performance
improvement, two new post-processing techniques, called post-processing
via variance normalization (PVN) and post-processing via dynamic
embedding (PDE), are proposed in this work. The PVN method normalizes
the variance of principal components of word vectors, while the PDE
method learns orthogonal latent variables from ordered input sequences.
The PVN and the PDE methods can be integrated to achieve better
performance. We apply these post-processing techniques to several popular
word embedding methods to yield their
post-processed representations.  Extensive experiments are conducted to
demonstrate the effectiveness of the proposed post-processing
techniques. 
\end{abstract}
\begin{keywords}
Word Representation, Variance Normalization, Sequential Extraction, Language processing
\end{keywords}
\section{Introduction}
\label{sec:introduction}

By transferring prior knowledge from large unlabeled corpus,
one can embed words into high-dimensional vectors with both semantic and
syntactic meanings in their distributional representations.  The design
of effective word embedding methods has attracted the attention of
researchers in recent years because of their superior performance in
many downstream natural language processing (NLP) tasks, including
sentimental analysis \cite{shin2016lexicon}, information retrieval
\cite{schutze2008introduction} and
machine translation \cite{ding2017visualizing}. In this work, two new
post-processing techniques, called post-processing via variance
normalization (PVN) and post-processing via dynamic embedding (PDE), are
proposed for further performance improvement. 

PCA-based post-processing methods have been examined in various research
fields. In the word embedding field, it is observed that learned word
vectors usually share a large mean and several dominant principal
components, which prevents word embedding from being isotropic.  Word
vectors that are isotropically distributed (or uniformly distributed in
spatial angles) can be differentiated from each other more easily.  A
post-processing algorithm (PPA) was recently proposed in \cite{mu2017all} to exploit this property. That is, the mean and several dominant principal components are removed by
the PPA method.  On the other hand, their complete removal may not be
the best choice since they may still provide some useful information.
Instead of removing them, we propose a new post-processing technique by
normalizing the variance of embedded words and call it the PVN method
here.  The PVN method imposes constraints on dominant principal
components instead of erasing their contributions completely. 

Existing word embedding methods are primarily built upon the concept
that ``You shall know a word by the company it keeps."
\cite{firth1957synopsis}.  As a result, most of current word embedding
 methods are based on training samples of ``(word, context)". Most context-based word embedding methods do not differentiate the word order in sentences, meaning that, they ignore the relative distance between
the target and the context words in a chosen context window.
Intuitively, words that are closer in a sentence should have stronger
correlation. This has been verified in \cite{khandelwal2018sharp}. Thus,
it is promising to design a new word embedding method that not only
captures the context information but also models the dynamics in a word
sequence.

To achieve further performance improvement, we propose the second
post-processing technique, which is called the PDE method. Inspired by
dynamic principal component analysis (Dynamic-PCA) \cite{dong2018novel},
the PDE method projects existing word vectors into an orthogonal
subspace that captures the sequential information optimally under a
pre-defined language model. The PVN method and the PDE method can work
together to boost the overall performance.  Extensive experiments are
conducted to demonstrate the effectiveness of PVN/PDE post-processed
representations over their original ones.

\section{Highlighted Contributions}

Post-processing and dimensionality reduction techniques in word
embeddings have primarily been based on the principal component analysis
(PCA).  There is a long history in high dimensional correlated data
analysis using latent variable extraction, including PCA,
singular spectrum analysis (SSA) and canonical correlation analysis
(CCA). They are shown to be effective in various applications. Among
them, PCA is a widely used data-driven dimensionality reduction
technique as it maximizes the variance of extracted latent variables.
However, the conventional PCA focuses on static variance while ignoring
time dependence between data distributions. It demands additional work
in applying the PCA to dynamic data. 

It is pointed out in \cite{mu2017all} that
embedded words usually share a large mean and several dominant principal
components.  As a consequence, the distribution of embedded words are
not isotropic. Word vectors that are isotropically distributed (or
uniformly distributed in spatial angles) are more differentiable from
each other.  To make the embedding scheme more robust and alleviate the
hubness problem \cite{radovanovic2010existence}, they proposed to remove
dominant principal components of embedded words.  On the other hand,
some linguistic properties are still captured by these dominant principal
components. Instead of removing dominant principal components
completely, we propose a new post-processing technique by imposing
regularizations on principal components in this work. 

Recently, contextualized word embedding has gained attention since it
tackles the word meaning problem using the information from the whole
sentence \cite{peters2018deep}. It contains a bi-directional long
short-term memory (bi-LSTM) module to learn a language model whose
inputs are sequences. The performance of this model indicates that the
ordered information plays an important role in the context-dependent
representation, and it should be taken into consideration in the
design of context-independent word embedding methods.

There are three main contributions in this work.  We propose two new
post-processing techniques in Sec. \ref{sec:VN} and Sec. \ref{sec:DE},
respectively. Then, we apply the developed techniques to several popular
word embedding methods and generate their
post-processed representations. Extensive experiments are conducted over various baseline models including SGNS~\cite{mikolov2013distributed}, CBOW~\cite{mikolov2013distributed}, GloVe~\cite{pennington2014glove} and Dict2vec~\cite{tissier2017dict2vec} in
Sec. \ref{sec:experiments} to demonstrate the effectiveness of
post-processed representations over their original ones.

\section{Post-Processing via Variance Normalization}\label{sec:VN}

We modify the PPA method \cite{mu2017all} by regularizing the variances
of leading components at a similar level and call it the post-processing
algorithm with variance normalization (PVN). The PVN method is described
in Algorithm \ref{PVNalgorithm}, where $V$ denotes the vocabulary set.

%%%%%%%%%%%%%%%%%%%%%%%%%%%%%%%%%%%%%%%%%%%%%%%%%%%%%%%%%%
\begin{algorithm}[ht]
\caption{Post-Processing via Variance Normalization (PVN)}\label{PVNalgorithm}
\begin{algorithmic}
\STATE \textbf{Input:} Given word representations $v(w)$, $w \in V$, and
threshold parameter $d$.
\STATE \textbf{1.} Remove the mean of $ \{v(w), w\in V\}$ \\ 
$\mu \leftarrow \dfrac{1}{|V|} \sum_{w\in V} v(w)$ and $\tilde{v}(w)\leftarrow v(w)-\mu$.
\STATE \textbf{2.} Compute the first $d+1$ PCA components \\
$u_1,\cdots, u_{d+1} \leftarrow \mbox{PCA}({\tilde{v}(w), w\in V})$ 
\STATE \textbf{3.} Compute the standard deviation for the first $d+1$ PCA components \\
$\sigma_1,\cdots,\sigma_{d+1} \leftarrow \mbox{variances of}\ u_1, \cdots, u_{d+1}$,
\STATE \textbf{4.} Determine the new representation \\ 
$v^{'}(w) \leftarrow \tilde{v}(w)-\sum_{i=1}^{d}\dfrac{\sigma_i-\sigma_{d+1}}
{\sigma_i}(u_{i}^{T} \tilde v(w))u_{i}$
\STATE \textbf{Output:} Processed representations $v^{'}(w)$, $w \in V$.
\end{algorithmic}
\end{algorithm}
%%%%%%%%%%%%%%%%%%%%%%%%%%%%%%%%%%%%%%%%%%%%%%%%%%%%%%%%%%

In Step 4, $(u_{i}^{T} \tilde v(w))$ is the projection of $\tilde v(w)$
to the $i_{th}$ principal component. We multiply it by a ratio factor
$\dfrac{\sigma_i - \sigma_{d+1}}{\sigma_i}$ to constrain its variance.
Then, we project it back to the original bases and subtract it from the
mean-removed word vector.  

To compute the standard deviation of the $i_{th}$ principal component of
processed representation $v'(w)$, we project $v'(w)$ to bases $u_j$:
\begin{equation}\label{eq:verify}
\begin{aligned}
u_{j}^{T} v'(w) &= u_{j}^{T} \tilde{v}(w)-\sum_{i=1}^{d}\dfrac{\sigma_i-\sigma_{d+1}}
{\sigma_i}(u_{i}^{T} \tilde v(w))u_{j}^{T}u_{i}\\
&= u_{j}^{T} \tilde{v}(w)-\dfrac{\sigma_j-\sigma_{d+1}}
{\sigma_j}(u_{j}^{T} \tilde v(w))u_{j}^{T}u_{j} \\
&=\dfrac{\sigma_{d+1}}{\sigma_j} u_{j}^{T} \tilde{v}(w).
\end{aligned}
\end{equation}
Thus, the standard deviation of all post-processed $j_{th}$ principal
component, $1\leq j\leq d$, is equal to $\sigma_{d+1}$. Thus, all
variances of leading $d+1$ principal components will be normalized to
the same level by the PVN. This makes embedding vectors more evenly
distributed in all dimensions. 

The only hyper-parameter to tune is threshold parameter $d$. The optimal
setting may vary with different word embedding baselines. A good rule of
thumb is to choose $d \approx D/50$, where $D$ is the dimension of word
embeddings. Also, we can determine the dimension threshold, $d$, by
examining energy ratios of principal components. 

\section{Post-Processing via Dynamic Embedding}\label{sec:DE}

\subsection{Language Model}\label{subsec:language}

Our language model is a linear transformation that predicts the current
word given its ordered context words. For a sequence of words: $w_1,
w_2, w_3, \cdots, w_n$, the word embedding format is represented as
$v(w_1), v(w_2), v(w_3), \cdots, v(w_n)$. Two baseline models, SGNS and GloVe,
are considered. In other words, $v(w_i)$ is the embedded word of $w_i$
using one of these methods. Our objective is to maximize the conditional
probability
\begin{equation}\label{eq:prob}
p(w_i\mid w_{i-c}, \cdots, w_{i-1}, w_{i+1}, \cdots, w_{i+c}),
\end{equation} 
where $c$ is the context window size. As compared to other language
models that use tokens from the past \cite{peters2018deep}, we consider
the two-sided context as shown in Eq. (\ref{eq:prob}) since they are 
about equally important to the center word distribution in language
modeling. 

The linear language model can be
written as
\begin{equation}\label{eq:model}
\tilde{v}(w_i)=\sum_{i-c\leq j \leq i+c, j\neq i} b_j \tilde{v}(w_{j})+ \Delta,
\end{equation}
where $\tilde{v}(w)$ is the word embedding representation after the
latent variable transform to be discussed in Sec. \ref{subsec:latent}.
The term, $\Delta$, is used to represent the information loss that
cannot be well modeled in the linear model. We treat $\Delta$ as a
neglible term and Eq. (\ref{eq:model}) as a linear approximation 
of the original language model.

\subsection{Dynamic Latent Variable Extraction}\label{subsec:latent}

We apply the dynamic latent variable technique to the word embedding
problem in this subsection. To begin with, we define an 
objective function to extract the dynamic latent variables. 
The word sequence data is denoted by 
\begin{equation}
{\bf W} = [w_1, w_2, \cdots , w_n],
\end{equation}
and the data matrix, ${\bf W}_i$, derived from ${\bf W}$ is formed using
the chosen context window size and its word embedding representation
${\bf V}_i$ from data ${\bf W}_i$:
\begin{eqnarray}
%\begin{aligned}
{\bf W}_i&=&[w_{i-c}, \cdots , w_{i-1}, w_{i+1}, \cdots, w_{i+c}] 
\in R^{1 \times 2c}, \\
{\bf V}_i&=&[v(w_{i-c}), \cdots ,
\quad v(w_{i-1}), v(w_{i+1}), \cdots,\\&& v(w_{i+c})]\in R^{D \times 2c},
%\end{aligned}
\end{eqnarray}
where $D$ is the word embedding dimension.  Then, the objective function
used to extract the dynamic latent variable can be written as
\begin{equation}\label{eq:optimization}
\begin{aligned}
\max_{{\bf A}, {\bf b}}\  \sum_{i} < {\bf A}^T {\bf V}_i {\bf b},\  {\bf A}^T v(w_i)> 
\end{aligned}
\end{equation}
where ${\bf A}$ is a matrix of dimension $D \times k$, ${\bf b} \in
R^{2c}$ is a weighted sum of context word representations, and
where $k$ is the selected dynamic dimension. 

We interpret ${\bf A}$ in Eq. (\ref{eq:optimization}) as a matrix that
stores dynamic latent variables.  If ${\bf A}$ contains all learned
dynamic principal components of dimension $k$, ${\bf A}^T {\bf V}_i$ is
the projection to dynamic principal components from all context word
representation and ${\bf A}^T v(w_i)$ is the projection of the center
word representation $v(w_i)$. Vector ${\bf b}$ is a weighted sum of context
representations used for prediction. We seek optimal ${\bf A}$ and ${\bf b}$
to maximize the sum of all inner products of predicted center word
representation ${\bf A}^T {\bf V}_i {\bf b}$ and the $i$th center word
representation ${\bf A}^T v(w_i)$.  The choice of the inner product
rather than other distance measures is to maximize the variance over
extracted dynamic latent variables. For further details, we refer to
\cite{dong2018novel}. 

\subsection{Optimization}

There is no analytical solution to Eq. (\ref{eq:optimization}) since
${\bf A}$ and ${\bf b}$ are coupled \cite{chen2002line}.  Besides, we
need to impose constraints on ${\bf A}$ and ${\bf b}$. That is, the
columns of ${\bf A}$ are orthonormal vectors while $||{\bf b}||=1$. The
orthogonality constraint on matrix ${\bf A}$ plays an important role.
For example, the orthogonality constraint is introduced for bilingual
word embedding for several reasons. The original word embedding space is
invariant and self-consistent under the orthogonal principal
\cite{artetxe2016learning,artetxe2018generalizing}. Moreover, without
such a constraint, the learned dynamic latent variables has to be
extracted iteratively, which is time consuming. Furthermore, the
extracted latent variables tend to be close to each other with a small
angle. 

We adopt the optimization procedure in Algorithm \ref{algorithm2} to
solve the optimization problem.  Note that parameter $\beta$ is used to
control the orthogonality-wise convergence rate. 

%%%%%%%%%%%%%%%%%%%%%%%%%%%%%%%%%%%%%%%%%%%%%%%%%%%%%%%%%%
\begin{algorithm}
\caption{Optimization for extracting dynamic latent variables}
\begin{algorithmic}
\STATE Initialize ${\bf A}$ and ${\bf b}$ randomly as learnable parameters.
\FOR{Training batch ${\bf W}_i $=1,...,m}
\STATE Find corresponding word embedding ${\bf V}_i$.
\STATE Predict the center word $\hat{v}(w_i)={\bf A}^T {\bf V}_i {\bf b}$.
\STATE Extract negative samples based on word frequency.
\STATE Update ${\bf A}$ and ${\bf b}$ by gradient descent to optimize Eq. (\ref{eq:objective}).
\STATE ${\bf b}:= {\bf b}/||{\bf b}||$
\STATE ${\bf A}:=(1+\beta){\bf A}-\beta {\bf A}{\bf A}^T {\bf A}$
\ENDFOR
\end{algorithmic}
\label{algorithm2}
\end{algorithm}
%%%%%%%%%%%%%%%%%%%%%%%%%%%%%%%%%%%%%%%%%%%%%%%%%%%%%%%%%%

We maximize the inner product over all tokens as shown in Eq.
(\ref{eq:optimization}) in theory, yet we adopt negative sampling for
parameter update in practice to save the computation. The objective
function can be rewritten as
$$
\max_{{\bf A}, {\bf b}} F({\bf A}, {\bf b}),
$$
where
\begin{equation}\label{eq:objective}
\begin{aligned}
F({\bf A}, {\bf b}) 
= \Big( \log\sigma(\left[({\bf A}^T {\bf V}_i {\bf b})^T {\bf A}^T v(w_i) \right])+ \\
\sum_{n=1}^{N}\mathbb{E}_{w_n\sim P_n(w)} \left[\log\sigma(-\left[({\bf A}^T 
{\bf V}_i{\bf b})^T {\bf A}^T v(w_n)\right])\right] \Big),
\end{aligned}
\end{equation}
and where $\sigma(x)=1/(1+exp(-x))$, $N$ is the amount of the negative
samples used per positive training sample, and negative samples $v(w_n)$
are sampled based on their overall frequency distribution. 

The final word embedding vector is a concatenation of two parts: 1) the
dynamic dimensions by projecting $v(w_i)$ to the learned dynamic subspace
${\bf A}$ in form of ${\bf A}^{T} v(w_i) $, and 2) static dimensions
obtained from static PCA dimension reduction in form of $PCA(v(w_i))$.

\section{Experiments}\label{sec:experiments}

\subsection{Baselines and Hyper-parameter Settings}

We conduct the PVN on top of several baseline models. For all SGNS related experiments, wiki2010 corpus\footnote{http://nlp.stanford.edu/data/WestburyLab.wikicorp.201004.txt.bz2}
(around 6G) is used for training. The vocabulary set
contains 830k vocabularies, which means words occur more than 10 times
are included. For CBOW, GloVe and Dict2vec, we all adopt the official released code and trained on the same dataset as SGNS. Here we set $d$=11 across all experiments.

For the PDE method, we obtain training pairs from the wiki2010 corpus to
maintain consistency with the SGNS model.  The vocabulary size is 800k.
Words with low frequency are assigned to the same word vectors for
simplicity. 

%%%%%%%%%%%%%%%%%%%%%%%%%%%%%%%%%%%%%%%%%%%%%%%%%%%%%%%%%%%%%%%%%%%
\begin{table}[htb]
\centering
\begin{tabular}{|l|l|l|}
\hline
\multicolumn{1}{|c|}{Name} & \multicolumn{1}{c|}{Pairs} & Year \\ \hline
WS-353                     & 353                        & 2002      \\ \hline
WS-353-SIM                 & 203                        & 2009      \\ \hline
WS-353-REL                 & 252                        & 2009      \\ \hline
Rare-Word                  & 2034                       & 2013      \\ \hline
MEN                        & 3000                       & 2012      \\ \hline
MTurk-287                  & 287                        & 2011      \\ \hline
MTurk-771                  & 771                        & 2012      \\ \hline
SimLex-999                 & 999                        & 2014      \\ \hline
Verb-143                   & 143                        & 2014      
\\ \hline
SimVerb-3500                   & 3500                        & 2016      
\\ \hline
\end{tabular}
\caption{Word similarity datasets used in our experiments, where pairs 
indicate the number of word pairs in each dataset.}\label{table1}
\end{table}
%%%%%%%%%%%%%%%%%%%%%%%%%%%%%%%%%%%%%%%%%%%%%%%%%%%%%%%%%%%%%%%%%%%

\subsection{Datasets}\label{subsec:datasets}

We consider two popular intrinsic evaluation benchmarks in our
evaluation: 1) word similarity and 2) word analogy. Detailed introduction can be found in \cite{wang2019evaluating}. Our proposed
post-processing methods work well in both evaluation methods as reported
in Sec. \ref{subsec:PVN} and Sec. \ref{subsec:PDE}.

%%%%%%%%%%%%%%%%%%%%%%%%%%%%%%%%%%%%%%%%%%%%%%%%%%%%%%%%%%%%%%%%%%%
\begin{table}[thb]
\center
\begin{tabular}{l|c|c|c}
\hline
\multicolumn{1}{c|}{Type}  & SGNS & PPA & PVN(ours)   \\
\hline
WS-353 & 65.7 & 67.6 & \textbf{68.1} \\
\hline
WS-353-SIM & 73.2 & 73.8 & \textbf{73.9} \\
\hline
WS-353-REL & 58.1 & 59.4 & \textbf{60.7} \\ 
\hline
Rare-Word & 39.5 & 42.4 & \textbf{42.9} \\
\hline
MEN & 70.2 & 72.5 & \textbf{73.2} \\ 
\hline
MTurk-287 & 62.8 & 64.7 & \textbf{66.4} \\
\hline
MTurk-771 & 64.6 & 66.2 & \textbf{66.8} \\
\hline
SimLex-999 & 41.6 & 42.6 & \textbf{42.8} \\
\hline
Verb-143 & 35.0 & 38.9 & \textbf{39.5} \\
\hline
SimVerb-3500 & 26.5 & 28.1 & \textbf{28.5} \\
\hline
Average & 47.8 & 49.8 & \textbf{50.3} \\ 
\hline
\end{tabular}
\caption{The SRCC performance comparison ($\times 100$) for SGNS alone, SGNS+PPA and SGNS+PVN against word similarity datasets, where
the last row is the average performance weighted by the pair number of
each dataset.} \label{table2}
\end{table}
%%%%%%%%%%%%%%%%%%%%%%%%%%%%%%%%%%%%%%%%%%%%%%%%%%%%%%%%%%%%%%%%%%%

%%%%%%%%%%%%%%%%%%%%%%%%%%%%%%%%%%%%%%%%%%%%%%%%%%%%%%%%%%%%%%%%%%%
\begin{table}[thb]
\center
\begin{tabular}{c|l|c|c|c}
\hline
\multicolumn{2}{c|}{Type:} & SGNS & PPA & PVN(ours)\\ 
\hline
\multirow{2}{*}{Google} & Add & 59.6 & 61.3 & \textbf{62.1} \\
\cline{2-5} 
 & Mul & 61.2 & 60.3 & \textbf{61.9} \\
\hline
\multirow{2}{*}{Semantic}  & Add & 57.8 & \textbf{62.4} & \textbf{62.4} \\
\cline{2-5} 
 & Mul & 59.3 & 59.5 & \textbf{60.9} \\
\hline
\multirow{2}{*}{Syntactic} & Add & 61.1 & 60.5 & \textbf{61.8} \\
\cline{2-5} 
 & Mul & 62.7 & 61.0 & \textbf{62.7} \\
 \hline
\multirow{2}{*}{MSR} & Add & 51.0 & 53.0 & \textbf{53.4} \\
\cline{2-5}
 & Mul & 53.3 & 53.3 & \textbf{54.9} \\
\hline
\end{tabular}
\caption{The SRCC performance comparison ($\times 100$) for SGNS alone, SGNS+PPA and SGNS+PVN against word analogy datasets.}\label{table3}
\end{table}
%%%%%%%%%%%%%%%%%%%%%%%%%%%%%%%%%%%%%%%%%%%%%%%%%%%%%%%%%%%%%%%%%%%

\subsubsection{Word Similarity}

Word similarity evaluation is widely used in evaluating word embedding
quality. It focuses on the semantic meaning of words. Here, we use the cosine distance measure and Spearman's rank correlation coefficients (SRCC) \cite{spearman1904proof} to
measure the distance and evaluate the similarity between our results and
human scores, respectively. We conduct tests on 10 popular datasets (see
Table \ref{table1}) to avoid evaluation occasionality.  For more
information of each dataset, we refer to the
website\footnote{http://www.wordvectors.org/}. 

\subsubsection{Word Analogy}

Due to the limitation of performance comparison in terms of word
similarity \cite{faruqui2016problems}, performance comparison in word
analogy is adopted as a complementary tool to evaluate the quality of
word embedding methods. Both addition and multiplication operations
are implemented to predict word $d$ here. In PDE method, we report commonly used addition
operation for simplicity.

We choose two major datasets for word analogy evaluation. They are: 1)
the Google dataset \cite{mikolov2013efficient} and 2) the MSR dataset
\cite{mikolov2013linguistic}. The Google dataset contains 19,544
questions. They belong to two major categories: semantic and
morpho-syntactic, each of which contains 8,869 and 10,675 questions,
respectively. We also report the results conducted on two Google
subsets.  The MSR dataset contains 8,000 analogy questions.
Out-of-vocabulary words were removed from both
datasets.\footnote{Out-of-vocabulary words are those appear less than 10
times in the wiki2010 dataset.}

\subsubsection{Extrinsic Evaluation}
For PVN method, we conduct further experiments over extrinsic evaluation tasks including sentiment analysis and neural machine translation (NMT). For both tasks, Bidirection LSTM is utilized as the inference tool. Two sentiment analysis dataset is utilized: Internet Movie Database (IMDb) and Sentiment Treebank dataset (SST). Europarl v8 dataset for english-french translation is utilized in our neural machine translation task. We report accuracy for IMDb and SST dataset and validation accuracy for NMT.

%%%%%%%%%%%%%%%%%%%%%%%%%%%%%%%%%%%%%%%%%%%%%%%%%%%%%%%%%%%%%%%%%%%
\begin{table}[thb]
\center
\begin{tabular}{c|c|c|c}
\hline
\multicolumn{1}{c|}{Baselines}  & IMDb & SST & NMT \\ \hline
\multicolumn{1}{l|}{SGNS}       & 80.92/\textbf{86.03} & 66.00/\textbf{66.76} & 50.50/\textbf{50.62}  \\ \hline
\multicolumn{1}{l|}{CBOW}       & 85.20/\textbf{85.81} & \textbf{67.12}/66.94 & 49.78/\textbf{49.97}  \\ \hline
\multicolumn{1}{l|}{GloVe}      & 83.51/\textbf{84.88} & 64.53/\textbf{67.74} & 50.31/\textbf{50.58}  \\ \hline
\multicolumn{1}{l|}{Dict2vec}   & 80.62/\textbf{84.40} & 65.06/\textbf{66.89} & 50.45/\textbf{50.56}  \\ \hline 
\end{tabular}
\caption{Extrinsic Evaluation for SGNS alone and SGNS+PVN. The first value is from orignal model while second value is from our post-processing embedding model.}
\label{PVN_Extrinsic}
\end{table}
%%%%%%%%%%%%%%%%%%%%%%%%%%%%%%%%%%%%%%%%%%%%%%%%%%%%%%%%%%%%%%%%%%%

\subsection{Performance Evaluation of PVN}\label{subsec:PVN}

The performance of the PVN as a post-processing tool for the SGNS baseline methods is given in Tables \ref{table3}. It also shows results of the
baselines and the baseline+PPA \cite{mu2017all} for performance
bench-marking. 

Table \ref{table2} compares the SRCC scores of the SGNS alone, the
SGNS+PPA and the SGNS+PVN against word similarity datasets.  We see that
the SGNS+PVN performs better than the SGNS+PPA. We observe the largest performance gain of the
SGNS+PVN reaches 5.2\% in the avarage
SRCC scores. It is also interesting to point out that the PVN is more
robust than the PPA with different settings in $d$. 

Table \ref{table3} compares the SRCC scores of SGNS,
SGNS+PPA and SGNS+PVN against word analogy datasets. We use
addition as well as multiplication evaluation methods. PVN
performs better than PPA in both. For the multiplication evaluation, the
performance of PPA is worse than the baseline.  In contrast, the
proposed PVN method has no negative effect as it performs consistently
well. This can be explained below.  When the multiplication evaluation
is adopted, the missing dimensions of the PPA influence the relative
angles of vectors a lot. This is further verified by the fact that some
linguistic properties are captured by these high-variance dimensions and
their total elimination is sub-optimal. 

Table \ref{PVN_Extrinsic} indicates extrinsic evaluation results. We can see that our PVN post-processing method performs much better compared with the original result in various downstream tasks.

%%%%%%%%%%%%%%%%%%%%%%%%%%%%%%%%%%%%%%%%%%%%%%%%%%%%%%%%%%%%%%%%%%%
\begin{table}[thb]
\center
\begin{tabular}{c|c|c|c}
\hline
\multicolumn{2}{c|}{Type}       & SGNS    & PVN(ours)      \\ \hline
\multicolumn{2}{l|}{WS-353}     & 65.7    & \textbf{65.9}  \\ \hline
\multicolumn{2}{l|}{WS-353-SIM} & 73.2    & \textbf{73.6}  \\ \hline
\multicolumn{2}{l|}{WS-353-REL} & 58.1    & \textbf{59.3}  \\ \hline
\multicolumn{2}{l|}{Rare-Word}  & \textbf{39.5}    & 38.6  \\ \hline
\multicolumn{2}{l|}{Google}     & 59.6    & \textbf{60.8}  \\ \hline 
\multicolumn{2}{l|}{Semantic}   & 57.8    & \textbf{59.6}  \\ \hline
\multicolumn{2}{l|}{Syntactic}  & 61.1    & \textbf{61.8}           \\ \hline
\multicolumn{2}{l|}{MSR}        & 51.0    & \textbf{51.6}           \\ \hline
\end{tabular}
\caption{The SRCC performance comparison ($\times 100$) for SGNS alone
and SGNS+PDE against word similarity and analogy datasets.}
\label{table4}
\end{table}
%%%%%%%%%%%%%%%%%%%%%%%%%%%%%%%%%%%%%%%%%%%%%%%%%%%%%%%%%%%%%%%%%%%

\subsection{Performance Evaluation of PDE}\label{subsec:PDE}

We adopt the same setting in evaluating the PDE method such as the
window size, vocabulary size, number of negative samples,
training data, etc. when it is applied to SGNS baseline. The final word representation is
composed by two parts: $\tilde{v}(w) = [v_s(w)^T, v_d(w)^T]^T$, where
$v_s(w)$ is the static part obtained from dimension reduction using
PCA and $v_d(w)={\bf A}^{T}v(w)$ is the projection of $v(w)$ to dynamic
subspace ${\bf A}$. Here, we set the dimensions of $v_s(w)$ and $v_d(w)$
to 240 and 60, respectively. The SRCC performance comparison of SGNS
alone and SGNS+PDE against the word similarity and analogy datasets
are shown in Tables \ref{table4}.  By
adding the ordered information via PDE, we see that the quality of word
representations is improved in both evaluation tasks. 

%%%%%%%%%%%%%%%%%%%%%%%%%%%%%%%%%%%%%%%%%%%%%%%%%%%%%%%%%%%%%%%%%%%
\begin{table}[thb]
\center
\begin{tabular}{c|c|c|c}
\hline
\multicolumn{2}{c|}{Type}       & SGNS    & PVN(ours)      \\ \hline
\multicolumn{2}{l|}{WS-353}     & 65.7    & \textbf{69.0}  \\ \hline
\multicolumn{2}{l|}{WS-353-SIM} & 73.2    & \textbf{75.3}  \\ \hline
\multicolumn{2}{l|}{WS-353-REL} & 58.1    & \textbf{61.9}  \\ \hline
\multicolumn{2}{l|}{Verb-143}   & 35.0    & \textbf{44.1}  \\ \hline
\multicolumn{2}{l|}{Rare-Word}  & 39.5    & \textbf{42.5}  \\ \hline
\multicolumn{2}{l|}{Google}     & 59.6    & \textbf{62.8}  \\ \hline 
\multicolumn{2}{l|}{Semantic}   & 57.8    & \textbf{62.8}  \\ \hline
\multicolumn{2}{l|}{Syntactic}  & 61.1    & \textbf{62.8}  \\ \hline
\multicolumn{2}{l|}{MSR}        & 51.0    & \textbf{53.7}  \\ \hline
\end{tabular}
\caption{The SRCC performance comparison ($\times 100$) for SGNS alone
and SGNS+PVN/PDE model against word similarity and analogy datasets.}
\label{table5}
\end{table}
%%%%%%%%%%%%%%%%%%%%%%%%%%%%%%%%%%%%%%%%%%%%%%%%%%%%%%%%%%%%%%%%%%%

\subsection{Performance Evaluation of Integrated PVN/PDE}

We can integrate PVN and PDE together to improve their individual
performance. Since the PVN provides a better word embedding, it can help
PDE learn better. Furthermore, normalizing
variances for dominant principal components is beneficial since they
occupy too much energy and mask the contributions of remaining
components. On the other hand, components with very low variances may
contain much noise. They should be removed or replaced while the PDE can
be used to replace the noisy components. 

The SRCC performances of the baseline SGNS method and the
SGNS+PVN/PDE method for the word similarity and the word analogy tasks
are listed in Table \ref{table5}. Better results are obtained across all datasets. 
The improvement over the Verb-143 dataset has a high
ranking among all datasets with either joint PVN/PDE or PDE alone.  This
matches our expectation since the context order has more contribution
over verbs. 

\section{Conclusion and Future Work}

Two post-processing techniques, PVN and PDE, were proposed to improve
the quality of baseline word embedding methods in this work. The two
techniques can work independently or jointly. The effectiveness of these
techniques was demonstrated by both intrinsic and extrinsic evaluation tasks.  

We would like to study the PVN method by exploiting the correlation of
dimensions, and applying it to dimensionality reduction of word
representations in the near future.  Furthermore, we would like to apply
the dynamic embedding technique to both generic and/or domain-specific
word embedding methods with a limited amount of data. It is also desired
to consider its applicability to non-linear language models.

% References should be produced using the bibtex program from suitable
% BiBTeX files (here: strings, refs, manuals). The IEEEbib.bst bibliography
% style file from IEEE produces unsorted bibliography list.
% -------------------------------------------------------------------------
\bibliographystyle{IEEEbib}
\bibliography{icme2019template}

\end{document}